\documentclass[letterpaper, 10 pt, conference]{ieeeconf}  

\IEEEoverridecommandlockouts                              

\usepackage{graphics} 
\usepackage{epsfig} 
\usepackage{mathptmx} 
\usepackage{times} 
\usepackage{amsmath} 
\usepackage{amssymb}  
\usepackage{tabularx}
\usepackage{hyperref}

\newcolumntype{L}{>{\raggedright\arraybackslash}X}

\newcommand{\fig}[1]{Fig.~\ref{#1}}

\usepackage{color, soul}

\usepackage{xcolor}

 \title{\LARGE \bf
Learning Orientation Distributions for Object Pose Estimation
}

\author{Brian Okorn$^{1}$, Mengyun Xu$^{1}$, Martial Hebert$^{1}$, David Held$^{1}$
\thanks{*This work was supported by NASA NSTRF, United States Air Force and DARPA under Contract No. FA8750-18-C-0092, National Science Foundation under Grant No.  IIS-1849154, and LG Electronics}
\thanks{$^{1}$Brian Okorn, Mengyun Xu, David Held and Martial Hebert are with Robotics Institute at Carnegie Mellon University, 5000 Forbes Ave, Pittsburgh, PA USA;
        {\tt\small bokorn@andrew.cmu.edu}}%
}

\begin{document}

\maketitle
\thispagestyle{empty}
\pagestyle{empty}

\begin{abstract}
For robots to operate robustly in the real world, they should be aware of their uncertainty.  However, most methods for object pose estimation return a single point estimate of the object's pose. In this work, we propose two learned methods for estimating a distribution over an object's orientation. Our methods take into account both the inaccuracies in the pose estimation as well as the object symmetries. Our first method, which regresses from deep learned features to an isotropic Bingham distribution, gives the best performance for orientation distribution estimation for non-symmetric objects.  Our second method learns to compare deep features and generates a non-parameteric histogram distribution.  This method gives the best performance on objects with unknown symmetries, accurately modeling both symmetric and non-symmetric objects, without any requirement of symmetry annotation. We show that both of these methods can be used to augment an existing pose estimator.  Our evaluation compares our methods to a large number of baseline approaches  for uncertainty estimation across a variety of different types of objects.
Code available at \href{https://bokorn.github.io/orientation-distributions/}{https://bokorn.github.io/orientation-distributions/}


\end{abstract}





\section{Introduction}
\label{sec:introcuction}
Pose estimation is a commonly used primitive in many robotic tasks such as grasping~\cite{kim2016planning}, motion planning~\cite{dantam2016incremental}, and object manipulation~\cite{thomas2018learning}. For grasping, pose estimation is regularly used to register an observed object to a 3D model for which grasp positions have been annotated~\cite{goldfeder2009columbia, ciocarlie2014towards}. In motion planning, many algorithms require the poses of objects in the environment, either for avoiding collisions~\cite{zucker2013chomp} or as a state representation used for planning how to manipulate the objects~\cite{dantam2016incremental}.
  
Most prior methods for pose estimation output a single best guess of each object's pose \cite{xiang2017posecnn,wang2019densefusion,brachmann2014learning,kehl2017ssd}. In contrast, for many robotic applications, we believe that it is important for a robot to be aware of the uncertainty underlying these estimates before taking an action. This uncertainty can be caused by environmental factors, such as occlusions, poor lighting, or object symmetry, or by  biases in the algorithm, induced by insufficient training sets. These factors can cause ambiguity with respect to the object's orientation. If this uncertainty is not taken into account, then the actions of the robot may cause irreversible damage to itself or its environment. For example, a poorly estimated pose estimate can cause a robot to knock over fragile objects while attempting to grasp them. In such cases, rather than taking potentially dangerous actions, the robot should instead capture more information about the environment in an attempt to reduce this uncertainty. Additionally, estimates of uncertainty allow the robot to fuse multiple estimates, through tracking, to achieve a more robust final pose estimate. Thus, methods for pose estimation for robotics should output a distribution of poses rather than just a single pose estimate.

\begin{figure}[t]
 	\centering
 	\includegraphics[width=0.48\textwidth]{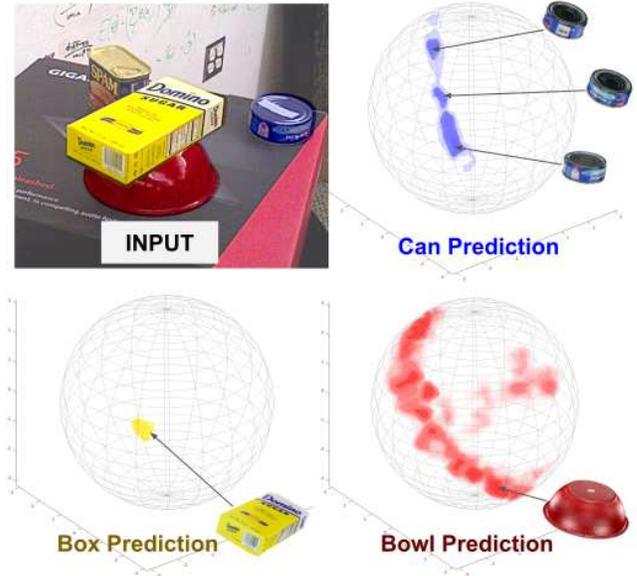}
 	\caption{Multi-modal distributions estimated by our Learned Comparison Histogram approach. These distributions are generated for the tuna can, bowl and sugar box using PoseCNN featurizations of the top right image. Here we see the estimator capturing multiple possible viewpoint for the tuna can, while still placing most of the probability density on the correct mode. It is also able to capture the full symmetry of the bowl without any symmetry labeling. In the case of unambiguous poses, like the sugar box, it is still capable of producing tight uni-modal distributions.}
 	\label{fig:comp_dist}
 	\vspace{-2em}
\end{figure}

We propose two novel methods for estimating orientation distributions.  The first method learns a uni-modal, parametric distribution in the form of an isotropic Bingham, regressed from deep learned features. This model is ideal for objects that are known to be non-symmetric. The second learns to estimate a multi-modal non-parametric distribution, in the form of a histogram distribution,  obtained using a learned comparison function over deep learned features.  We find that this second method works well for objects with unknown symmetries, accurately modeling both symmetric and non-symmetric objects, without any requirement of symmetry annotation.


We compare our learned methods against other statistically driven methods for estimating parametric and non-parametric orientation distributions.  We test each method on the pre-trained feature representations from two state-of-the-art pose estimation methods~\cite{xiang2017posecnn,wang2019densefusion}, and evaluate on a large pose estimation dataset~\cite{xiang2017posecnn} that has been used in a number of recent works~\cite{wang2019densefusion,tremblay2018deep}.

\section{Related Work}
\label{sec:relatedwork}
\subsection{Pose Estimation} 
Previous methods for pose estimation fall into four major categories: segmentation based methods, local coordinate based methods, image template based methods, and direct regression methods. Segmentation based algorithms~\cite{wong2017segicp,zeng2017multi} use an object segmentation algorithm to isolate the points associated with the target object. The segmented depth pixels can be registered with a 3D model of the object using Iterative Closest Point (ICP) algorithms. Local coordinate methods densely predict the 3D location of each pixel with respect to the original object model~\cite{brachmann2014learning}. These local coordinates define correspondences between the model and the image pixel locations; which are then used with RANSAC~\cite{fischler1981random} to find the object's pose. Alternatively, instead of densely estimating coordinates, the coordinates of an object's bounding box can be regressed can be regressed~\cite{tremblay2018deep}. Image template methods~\cite{hinterstoisser2012model,cao2016real,hodavn2015detection} render a template image at multiple viewpoints around the object model and compute a feature representation at each pose. The objects pose is estimated by looking up the nearest object templates, either by successive pruning of candidates~\cite{hinterstoisser2012model}, a hashing function~\cite{hodavn2015detection,drost2010model}, or by GPU parallelized comparison~\cite{cao2016real}. These coarse estimates tend to be refined using ICP. Recently, deep learned methods have been explored, which can directly regress the object's pose using RGB images~\cite{xiang2017posecnn} or densely fused image and point features~\cite{wang2019densefusion}. Additionally, learned latent spaces have been explored as object pose representations~\cite{sundermeyer2018augment,wohlhart2015learning,balntas2017pose}. In this work, we focus not on improving the accuracy of the underlying pose estimate but in adding a model of the estimates uncertainty over the entire orientation space.

\subsection{Pose Distribution Estimation}
While most prior methods for pose estimation output a single best guess of each object's pose, there has been some recent work on estimating pose distributions. Su~\cite{su2015render} estimated uncertainty distributions over the individual camera view angles relative to classes of objects through a soft classification method. Marton~\cite{marton2018improving} estimated a conditional probability distribution over orientations, in the form of a confusion matrix generated over rendered point clouds. Glover~\cite{glover2012monte} fit mixtures of Bingham distributions to clusters of local point cloud features to estimate an orientation distribution. Similarly, Riedel~\cite{riedel2016multi} combined multiple pose estimates using Bingham mixture models. However, unlike this work, they do not evaluate uncertainty estimation with respect to existing deep learned methods or with respect to log likelihood.

Other previous work has estimated a distribution over the object coordinates~\cite{brachmann2016uncertainty} or bounding box coordinates~\cite{tremblay2018deep}. However, these methods do not output a distribution over poses, nor do they evaluate whether the distributions themselves are reasonable. One previous paper evaluates distributions over the poses of object classes~\cite{su2015render}, mostly focusing on azimuth estimation.  In contrast, we estimate the orientation distribution of specific object instances and over the full space of orientation.

Most recently, Deng~\cite{Fox-RSS-19} used a learned feature space to estimate multimodal uncertainty distributions over rotations, and used those estimates for particle filter tracking. However, this work did not quantitatively evaluate the uncertainty distribution itself, nor did it compare to other approaches for estimating orientation distributions. Additionally, this method requires the use of a specifically learned autoencoder representation~\cite{sundermeyer2018augment}. Manhardt~\cite{manhardt2019explaining} explored learning orientation distributions through PCA analysis of multiple orientation hypotheses, trained using a winner-take-all approach. While this method does visualize their distributions as Bingham distributions, they do not investigate the accuracy of the underlying uncertainty distribution beyond qualitative analysis.

\subsection{Neural Network Uncertainty Estimation}
Because deep learning is a popular method for many computer vision tasks (including pose estimation), many approaches have explored how to estimate uncertainty from neural networks.  The most popular approaches include Monte Carlo Dropout~\cite{gal2016dropout} to estimate epistemic uncertainty, and regressing to the parameters of a distribution~\cite{kendall2017uncertainties} to estimate aleatoric uncertainty. We evaluate both of these approaches in this work. 

\subsection{Pose Tracking}
Tracking 6D rotation has been done using Kalman filters over Bingham Distributions~\cite{srivatsan2017bingham, gilitschenski2015unscented}. Bingham distributions~\cite{bingham1974antipodally} are well suited for this problem when the orientation distribution is expected to be unimodal, as they well model rotation quaternion and their composition is well defined. Additionally, particle filtering~\cite{Fox-RSS-19, grossmann2017fast} as well as histogram filtering~\cite{marton2018improving} have been used to sequentially improve and track object pose. The distribution estimates estimated by our method can be similarly used to improve pose estimate accuracy.
\section{Background}
\subsection{Orientation Representation}
Unit quaternions are used as our rotation representation, as they are a compact, numerically stable representation that does not suffer from singularities or gimbal lock. For these reasons, they are the preferred representation of 3D orientation in many papers for both robotics and deep learning \cite{xiang2017posecnn, wang2019densefusion}. Additionally, unit quaternions have well studied parametric distributions, as well as several uniform sampling strategies \cite{shoemake1992uniform, yershova2010generating, perez2013uniform}. For more background on quaternions, we refer the reader to \cite{dam1998quaternions}.

\subsection{Bingham distributions}
\label{sec:distribution_representations}

One of our proposed methods, described in Section~\ref{sec:bingham_regression}, makes use of a Bingham distribution~\cite{bingham1974antipodally}. A Bingham distribution is an antipodal distribution over the surface of a sphere, equivalent to a Gaussian distribution conditioned to lie on the orientation space, SO(3). Bingham distributions have been used for both orientation tracking and filtering~\cite{srivatsan2017bingham,glover2012monte,riedel2016multi}. These distributions are parameterized by an orthogonal 4x4 quaternion rotation matrix $\mathbf{M}$, which describes how the distribution will be rotated on the 3-sphere, and the diagonal 4x4 concentration matrix $\mathbf{Z}$ which describes the spread of the distribution. Similar to Gaussian distributions, Bingham distributions can be simplified to an isotropic distribution, parameterized by a mean quaternion and a single concentration parameter, analogous to variance for a Gaussian) . 

\section{Methods for Estimating Orientation Distributions}
\label{sec:approach}

We introduce two novel algorithms for learning  orientation distributions. These methods can be used to augment many existing pose estimators, without decreasing the single point accuracy of the underlying system. In this work, we focus on estimating only the uncertainty of the object's orientation, and not its full 6D pose. However, given a distribution over the object's orientation, a distribution over translation can also be estimated using Rao-Blackwellized particle filter sampling~\cite{Fox-RSS-19}.

\begin{figure}[t]
 	\centering
 	\includegraphics[width=0.48\textwidth]{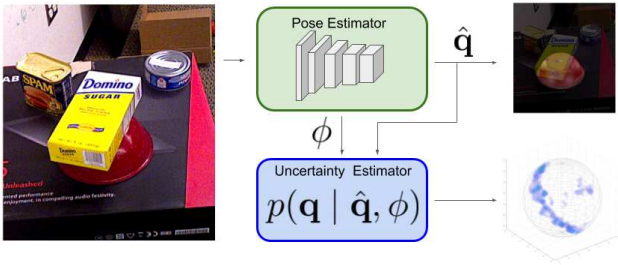}
 	\caption{System pipeline for estimating orientation distributions about an existing pose estimator. The base pose estimator generates an orientation $\hat{\mathbf{q}}$ and a featurization $\phi$ of the input, one or both of which are used to estimate a uncertainty distribution over possible poses. We render this distribution in as a heat map in axis angle space, lower right, with each orientation being plotted as point in the directions of the axis of rotation and at a distance away form the origin equal to the angle of rotation.}
 	\label{fig:overview}
 	\vspace{-2em}
\end{figure}




\subsection{\textbf{Bingham Distribution Regression}}
\label{sec:bingham_regression}
Our first method is designed to estimate the distribution of non-symmetric objects.  For such objects, we regress the parameters of a Bingham distribution from deep learned object features. 
Our method builds off of a base pose estimator which extracts a set of features $\phi(I)$ from a cropped image $I$ of the target object.  The base pose estimator then regresses from these features $\phi(I)$ to a single point estimate $\bar{\mathbf{q}}$ of the object's orientation. The focus of our approach is not in obtaining these features $\phi(I)$ or in learning the point estimate $\bar{\mathbf{q}}$; rather, these are provided as an input to our system.  We evaluate a couple of different options for feature extraction, as explained in Section \ref{sec:implementation_details}, and show that our method works for both.

We use the orientation $\bar{\mathbf{q}}$ as the mean of the Bingham distribution. From the features $\phi(I)$, our method learns to regress the remaining parameters of the Bingham distribution, explained below. The parameters of this method are learned by maximizing the log likelihood of the ground-truth pose for each image in the training set.



For simplicity, we limit our Bingham distribution to having an isotropic covariance, requiring only a single parameter $\sigma$ to be learned. The orthogonality constraint on $M$ can be handled using the Cayley’s factorization of the of 4D rotations~\cite{perez2017cayley}, giving us a parameterization of $\mathbf{M}$ into two unit norm quaternions, $\mathbf{q}_L$ and $\mathbf{q}_R$. By setting $\mathbf{q}_L=\bar{\mathbf{q}}$ and $\mathbf{q}_R$ to the identity quaternion, we both simplify the regression and guarantee that the distribution is centered about $\bar{\mathbf{q}}$. This parameterization can be used to regress an anisotropic Bingham, but we found that the isotropic Bingham produced more accurate results and a more stable training procedure. Results using the full Bingham regression are included as a baseline; see Section \ref{sec:full_bingham_regression} for details. 


\begin{figure}[b]
 	\centering
 	\includegraphics[width=0.48\textwidth]{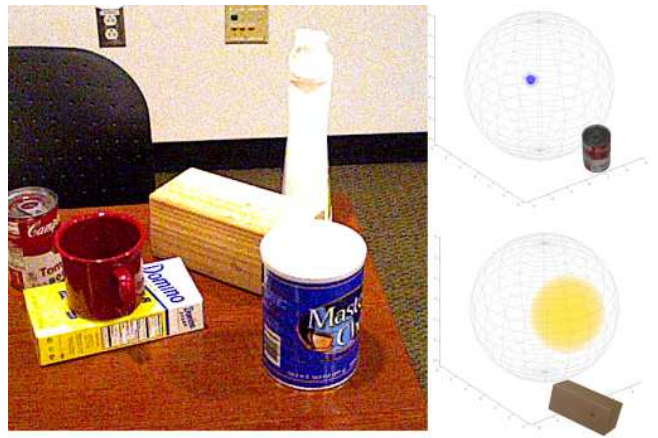}
 	\caption{Isotropic Bingham distributions regressed for the soup can, top, and the wood block, bottom, using DenseFusion featurization. The estimator is able to tightly fit a Bingham to the unambiguous pose of the soup can, but is not able to capture the multi-modal symmetry of the wood block. The only recourse is to inflate the uncertainty in an attempt to capture multiple modes.}
 	\label{fig:bing}
\end{figure}

\subsection{\textbf{Multi-modal Distribution Regression}}
\label{sec:multimodal_distribution_regression}
For symmetric objects, or objects that appear symmetric from certain poses or under particular occlusion patterns, a uni-modal Bingham distribution may not be sufficient to capture the object's uncertainty.  In such cases, a multi-modal histogram distribution may be more appropriate.

We use a $k$-nearest neighbor representation over a uniformly gridded space of unique orientations. In this work, we using the discretization method described by Straub~\cite{straub2016efficient}, as it enforces a near uniform distance between vertices, but any uniform sampling or gridding method could be used. The likelihood estimates at these vertices are interpolated using inverse distance weighting to the $k$ nearest orientations with respect to angular distance.  These interpolated values are normalized by dividing by the surface integral of the interpolation over the space of unique rotations, to form a valid continuous probability distribution.

A naive approach to obtaining such a histogram would be to regress from some latent features $\phi(I)$ directly to the parameters of a multi-modal histogram, $p(\mathbf{q} \mid \phi)$.  We include this approach as one of our baselines; see Section \ref{sec:hist_reg} for details.  We show that such a method leads to poor results, due to the inability of such a method to generalize to unseen object viewpoints.



\noindent \textbf{Learned Comparison Histogram:}
\label{sec:likelihood estimation}
We instead learn a comparison function $f(\phi(I_j) \mid \phi(I))$ between the features $\phi(I)$ and the features $\phi(I_j)$, which are computed from an image of the object rendered at orientation $\mathbf{q}_j$. To simplify notation, we will write this comparison function as $f(\phi_j \mid \phi)$ These rendered orientations are selected using the gridding described above. Our feature comparison function, once normalized, is specifically trained to approximate the posterior, e.g. $f(\phi_j, \phi) \approx \hat{p}(\mathbf{q}_j \mid \phi)$, as described below.

To mimic the posterior $\hat{p}(\mathbf{q}_j \mid \phi)$, we train the comparison function, $f(\phi_j \mid \phi)$, using an interpolated negative log likelihood loss. Specifically, given a ground-truth orientation of $\mathbf{q^*}$, we minimize the loss 
\begin{align}
\begin{split}
\mathcal{L}(\mathbf{q}^*; \phi) =& -\log\left(\sum_{k=1}^K \hat{p}(\mathbf{q}_k \mid \phi) / d(\mathbf{q}^*, \mathbf{q}_k)\right) \\
&  \quad + \log\left(\sum_{j=1}^N \hat{p}(\mathbf{q}_j \mid \phi)\right)\\
\label{eq:loss}
\end{split}
\end{align}

where $d(\mathbf{q}^*,\mathbf{q}_k)$ is the minimum angular distance between orientations $\mathbf{q}^*$ and $\mathbf{q}_k$.  The set $\{\mathbf{q}_1,\ldots,\mathbf{q}_K\}$ are the $K$ nearest gridded orientations to $\mathbf{q}^*$, and $\{\mathbf{q}_1,\ldots,\mathbf{q}_N\}$ are all of the orientations in our gridding. In our experiments, we use $K = 4$. 


We pre-compute the features $\phi_j$ using a rendered image, $I_j$, of the object generated with uniform lighting and no occlusions at orientation $\mathbf{q}_j$.  This image is then passed through the base pose estimator to extract features $\phi_j$. Note that, if the featurization $\phi(\cdot)$ is fixed, the features $\phi_j$ can be pre-computed and cached.  
This method is capable of learning tight uni-modal distributions when the pose of the object is unambiguous, like the sugar box in \fig{fig:comp_dist}, while still maintaining the flexibility to learn complicated multi-modal distribution cause by symmetry, as is the case with the bowl or ambiguity cause by similar viewpoints, as seen with the tuna can. 

Although the feature comparison function $f(\phi_j | \phi)$ can be parameterized in a variety of ways, we parameterize it as a neural network that takes concatenated features $\phi$ and $\phi_j$ as input. Implementation details of our specific architecture and training procedure can be found in Section \ref{sec:implementation_details}.

\section{Experimental Evaluation}

\subsection{Baselines}

We compare our method to other common distribution estimation approaches.While the set of methods we compare to is far from exhaustive, we believe it represents a good sampling of the major classes of distribution estimation algorithms.

\subsubsection{\textbf{Fixed Isotropic Bingham}}
\label{sec:fixed_bingham}
Given a base pose estimator (such as~\cite{wang2019densefusion,xiang2017posecnn}) which outputs a single point estimate $\bar{\mathbf{q}}$ of the object's orientation, a simple baseline method for estimating an orientation distribution is to fit a Bingham centered about $\bar{\mathbf{q}}$, with a fixed isotropic concentration parameter, $\sigma$. This parameter can be tuned independently for each object, using cross-validation. In our experiments, we fit this parameter using a sub-random search~\cite{bousquet2017critical} over a validation set, maximizing the log likelihood of the ground truth orientation.

Note that, unlike our method described in Section \ref{sec:approach}, the uncertainty of this baseline does not depend on the input image; rather, a single uncertainty parameter is used for all images of a given object type.  Thus, this approach is not sensitive to the uncertainties that can be induced by environmental factors such as lighting, viewpoint, or occlusions.  We show that this approach performs significantly worse than our method which outputs image-dependent uncertainty estimates.

\subsubsection{\textbf{Mixture of Isotropic Binghams}}
\label{sec:mixed_bingham}
Some methods, such as DenseFusion~\cite{wang2019densefusion}, output a set of orientation estimates $\mathbf{q}_i$, each with a corresponding confidence $c_i$.  A mixture of isotropic Bingham distributions can be fit to this output, with each isotropic Bingham distribution centered at the orientation estimate $\mathbf{q}_i$ with a fixed concentration parameter, $\sigma$, similarly tuned using cross-validation. These Bingham distributions are combined into a single mixture distribution by weighting each one by its confidence $c_i$, where the confidence scores are normalized to sum to one.


\subsubsection{\textbf{MC-Dropout Ensemble}}
\label{sec:dropout}
Monte Carlo Dropout~\cite{gal2016dropout} has been used to approximate the epistemic uncertainty of a network's predictions, using dropout to simulate an ensemble of estimators. PoseCNN~\cite{xiang2017posecnn} includes a dropout layer, whereas we retrained DenseFusion~\cite{wang2019densefusion} with an additional dropout layer inserted into the network. At test time, $n$ forward passes of the network are run on each observation, with dropout active, to generate $n$ orientation estimates for each input.  This process generates an estimate of the epistemic uncertainty and is mathematically equivalent to a deep Gaussian process~\cite{gal2016dropout}. We make the assumption that these samples are drawn from a Bingham distribution and fit the parameters of such a distribution to the sampled orientation estimates. The number of forward passes used provides a trade-off between the accuracy of the uncertainty estimates and the speed of computation; following previous work~\cite{kendall2015bayesian}, we choose $n=50$ as a balance between accuracy and speed.

\subsubsection{\textbf{Confusion Matrix}}
\label{sec:confusion_matrix}
As described in~\cite{marton2018improving}, a confusion matrix can be used to estimate the conditional uncertainty $p(\mathbf{q}^* \mid \hat{\mathbf{q}})$ of an estimate $\hat{\mathbf{q}}$.  The confusion matrix is computed over a discretization of the orientation space. This method counts how often the ground-truth orientation $\mathbf{q}^*$  is classified as $\hat{\mathbf{q}}$ by the our base estimator in a training or validation set. As with our method, we use the orientation discretization from Straub~\cite{straub2016efficient} to define the discretization of the confusion matrix. 

Specifically, we form a $n \times n$ matrix, $\mathbf{X}$, where $n$ is the number of orientations in our discretization. Each row represents the estimated poses $\hat{q}_j$, whereas each column represents the ground-truth poses $q^*$.  We initialize this matrix to 0. To compute the elements of this matrix, we iterate over our dataset.  For each image $I_j$, we compute an estimated orientation $\hat{q}_j$ with a base pose estimator (e.g.~\cite{xiang2017posecnn} or~\cite{wang2019densefusion}). Given the ground-truth pose $q^*$, we then increment the value of the matrix corresponding to the row and column of $(\hat{q}_j, q^*)$. A small constant $\epsilon$ is to each element of the confusion matrix for Laplace smoothing, and the rows are then normalized using the procedure described in Section \ref{sec:multimodal_distribution_regression}. 

At inference time, we compute the estimated orientation $\hat{\mathbf{q}}$ using the base estimator. The row in the confusion matrix that corresponds to this estimated orientation gives the estimated value of the distribution $p(\mathbf{q}^* \mid \hat{\mathbf{q}})$.

\subsubsection{\textbf{Full Bingham Regression}}
\label{sec:full_bingham_regression}
Using the parameterization described in Section \ref{sec:bingham_regression}, we can regress the parameters of a full Bingham distribution. We still require that the Bingham be centered at the output of the estimator, $\bar{\mathbf{q}}$, but the covariance can be dilated and rotated about this point. The four parameters of the diagonal concentration matrix, $\mathbf{Z}$, can be simplified to three parameters by subtracting the maximum value, without loss of generality \cite{bingham1974antipodally}. To rotate the distribution about $\bar{\mathbf{q}}$, the 4D rotation matrix $\mathbf{M}$, can be post-multiplied by the four dimensional rotation matrix $\mathbf{Q}$, using a three dimensional rotation $\mathbf{R}_P$ parameterized by the quaternion $\mathbf{q}_P$, $\mathbf{Q} = diag\left(\begin{bmatrix} 1 & \mathbf{R}_P \end{bmatrix} \right)$.


\subsubsection{\textbf{Direct Histogram Regression}}
\label{sec:hist_reg}
As mentioned previously, we test directly regressing from the features $\phi(I)$ to  the histogram values at each gridded orientation $\mathbf{q}_j$, as opposed to computing these values based on feature comparisons. For this baseline, the values at each grid cell, $p(\mathbf{q} \mid \phi)$, are estimated using a neural network, which receives as input the latent features $\phi$ and regresses an unnormalized posterior, $\hat{p}(\mathbf{q}_j \mid \phi)$. As before, we train this function with the log likelihood loss of equation~\ref{eq:loss}.  Also as before, we normalize over all of the gridded orientations, and use the gridding from Straub~\cite{straub2016efficient}.

\subsubsection{\textbf{Cosine Feature Difference}}

As an ablation of our learned comparison method from Section \ref{sec:multimodal_distribution_regression}, we evaluate using the cosine distance 
as the feature comparison function, e.g. $f(\phi_j, \phi) = \phi_{j} \cdot \phi / (||\phi_{j}|| \, ||\phi||)$. For this ablation, the cosine distance replaces our learned comparison function, to evaluate the benefits to learning such a comparison function. This distance function $f(\phi_j, \phi)$ is used to approximate $\hat{p}(\mathbf{q}_j \mid \phi)$ in the same manner as described in Section \ref{sec:multimodal_distribution_regression}.

\subsection{Dataset}

To evaluate the accuracy of our methods for uncertainty estimation as well as the baselines, we use the YCB Video dataset~\cite{xiang2017posecnn}, a commonly used pose estimation dataset. This dataset is comprised of videos of 21 objects in various cluttered tabletop scenes, with segmentation and 6D pose annotations. Each object in the dataset is accompanied by a textured mesh.  Among the 21 objects, four objects contain discrete rotational symmetries, meaning the objects have a rotational symmetry with respect to a discrete set of rotations.  One object (the bowl) has a continuous rotational symmetry, having a symmetric axis about which the object can be freely rotated. Twelve of the videos are held out as a test set, leaving 80 videos for training. We focus on this dataset for our evaluation, as the two base estimators that we evaluate, DenseFusion~\cite{wang2019densefusion} and PoseCNN~\cite{xiang2017posecnn}, have made the pretrained weight for these objects available.


\subsection{Implementation Details}
\label{sec:implementation_details}
We tested each method for estimating orientation distributions using both PoseCNN~\cite{xiang2017posecnn} and DenseFusion~\cite{wang2019densefusion} features. When generating features with DenseFusion, we used the segmentation estimated by PoseCNN for training images, as is done in the original publication~\cite{wang2019densefusion} and the ground truth segmentation for the rendered images used for our non-parametric distributions. We use the global feature produced by DenseFusion for our multi-modal methods, while the maximum confidence local feature is used in our Bingham Regression method. These were experimentally verified to produce the best results in each method. All features are generated using pretrained models without further fine-tuning.

For PoseCNN features, we use the output of the last hidden layer of the network's orientation head. When generating PoseCNN features for rendered images, it is possible for the estimator to not detect the target object, as the network jointly estimates a segmentation mask as well as the pose of the object. In these cases, we evaluated each method using the featurization of the detected object whose mask maximally overlaps the target object. When the estimator failed to find any object in an image, we set the feature to the zero vector. This process is only used for rendered images. For real images, only the features of objects detected by PoseCNN are used.

Our methods are trained using a combination of real and rendered data. This data is resampled to ensure a uniform coverage over SO(3) using the discretization method described in Section~\ref{sec:distribution_representations}. In this case, we use coarser discretization than our distribution gridding, with a maximum distance to the nearest bin center of about 26 degrees.

Our non-parametric methods used a simple three layer neural network with 4096 neurons on each hidden layer, dropout and ReLU activations on the input and first hidden layer, and sigmoid activation on the output. The parametric methods draw inspiration from DenseFusion~\cite{wang2019densefusion}, using four fully connected layers, with 640, 256, and 128 neurons on the hidden layers and ReLU activation functions. 

\subsection{Evaluation Method}

We evaluate each orientation distribution estimator on each example in the YCB test set and record the log likelihood of the ground-truth pose, clipped to a minimum of 1e-6. A likelihood distribution is computed for each of these images and the likelihood of the ground truth pose is computed given that distribution. For multi-modal methods, the interpolation described in Section \ref{sec:multimodal_distribution_regression} is used, while Bingham based methods use standard Bingham likelihood. The log likelihood evaluation metric allows us to evaluate whether the distribution is correctly placing probability mass in the appropriate locations. 

\begin{table*}[t]
\small
\centering
\begin{tabular}{|c||c|c||c|c|c|c|c|c|c|c|c|c|}
\hline
  & \multicolumn{2}{c||}{Our Method} & \multicolumn{7}{c|}{Baselines} \\
\hline
 & Bingham & Learned & Fixed & Bingham & & Confusion & Cosine & Full & Histogram \\
Objects & Regression & Comparison & Bingham & Mixture & Dropout & Matrix & Distance & Bingham & Regression \\
\hline
\multicolumn{9}{l}{Non-Symmetric} \\
\hline
DenseFusion 
 & \textbf{2.80} & 1.18 & 1.74 & 0.66 & 0.70 & 1.63 & -1.90 & 2.56 & 0.28 \\
PoseCNN 
 & 1.91 & 2.17 & 1.50 & - & \textit{2.71} & -2.46 & -0.92 & 1.95 & 1.87 \\
\hline
\multicolumn{9}{l}{Symmetric} \\
\hline
DenseFusion 
 & -3.81 & -5.54 & -3.66 & -2.27 & -8.09 & -2.91 & -2.23 & -4.18 & -2.57 \\
PoseCNN 
 & -8.82 & \textbf{-0.52} & -9.18 & - & -5.28 & -7.75 & -1.55 & -3.70 & \textit{-1.23} \\
\hline
\multicolumn{9}{l}{All} \\
\hline
DenseFusion 
 & \textit{1.72} & 0.08 & 0.86 & 0.18 & -0.74 & 0.88 & -1.95 & 1.46 & -0.19 \\
PoseCNN
 & 0.19 & \textbf{1.74} & -0.22 & - & 1.43 & -3.31 & -1.02 & 1.05 & 1.37 \\
\hline
\end{tabular}
\caption{Mean Log Likelihood of Ground Truth Orientation.  For each grouping, best-scoring methods are marked in bold; second-best scoring methods are indicated by italics. }
\label{tbl:loglik}
\vspace{-2em}
\end{table*}

\section{Results}

The log likelihood results of our method and all the baselines can be seen in Table~\ref{tbl:loglik}. We separate the objects into symmetric and non-symmetric objects and evaluate each method using DenseFusion~\cite{wang2019densefusion} and PoseCNN~\cite{xiang2017posecnn} features. We find that our method of isotropic Bingham regression performs the best for non-symmetric objects when combined with DenseFusion features. Good performance is also obtained with a Bingham distribution fit to samples from MC Dropout using PoseCNN features.  A uni-modal Bingham distribution is able of capture the orientation uncertainty of non-symmetric objects when the distribution is tightly fit around a mean orientation, as shown by the tomato soup can in \fig{fig:bing}.  However, such a method will struggle with symmetric objects, like the wooden block in \fig{fig:bing}, or objects that appear symmetric from particular views or under particular occlusion patterns. 

While the Full Bingham Regression performed similarly to the Isotropic Bingham Regression, we found this method to be less numerically stable in training, as it requires the gradients for the normalization constant of an anisotropic Bingham distribution. The gradients of the isotropic normalization constant, however, proved to be more stable and cause few problems in training. Our experiments demonstrate that this longer training time provides little benefit over the isotropic version. 

\begin{figure}[t]
 	\centering
 	\includegraphics[width=0.48\textwidth]{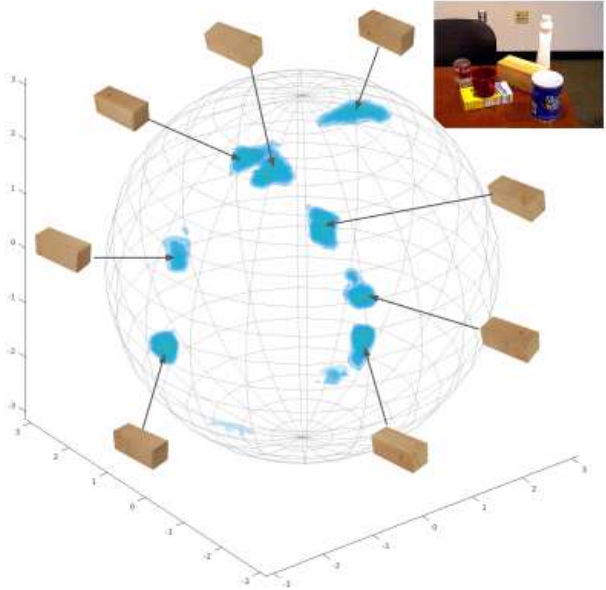}
 	\caption{ Multimodal distribution of the wood block's symmetries captured by the Learned Comparison Estimator, using PoseCNN features. There are eight distinct modes, associated with four 90 degree rotations about the long axis multiplied by two 180 degree rotations about one of the short axes. This distribution is impossible to well model with a single Bingham distribution, as shown in \fig{fig:bing}, but can be easily captured by a multi-modal histogram.}
 	\label{fig:block}
 	\vspace{-2em}
\end{figure}

For symmetric objects, Table~\ref{tbl:loglik} shows that learning a non-parametric histogram distribution is best able to capture the multi-modal nature of the uncertainty of such objects.  Specifically, Table~\ref{tbl:loglik} shows that our Learned Comparison Histogram estimation method has the best log likelihood, when using PoseCNN features. PoseCNN features using Histogram Regression is also among the top scoring methods for this task, although performance is significantly worse than our method. Note that the log likelihoods of the symmetric objects are expected to be lower than the log likelihood for non-symmetric objects, since the optimal distribution will spread the probability mass evenly over each symmetric mode, leading to a lower likelihood at each mode. This can be seen when our method correctly distributes the probability density to all eight of the wood block's symmetric modes, shown in \fig{fig:block}. Overall, our learned comparison based method for estimating a non-parametric distribution is best able to capture the uncertainty across the full set of objects, having the flexibility to model multi-modal distributions for objects with various types of symmetries, while still being able to concentrate the probability mass over a single mode when necessary.

We note that the log likelihood values in Table~\ref{tbl:loglik} may be hard for the reader to interpret directly; for reference, a uniform distribution, where every orientation is equally likely, would be expected to obtain a log likelihood of -2.29. As shown in Table~\ref{tbl:loglik},  some distributions perform worse than the uniform distribution. This is likely caused by overestimating the certainty of the output, i.e. the distribution for such methods is often concentrated around a single incorrect mode.  In such cases, the method fails to put sufficient probability mass in regions of the pose space far from this incorrect mode, leading to a very low log likelihood at the ground-truth pose.

Table~\ref{tbl:loglik} also reveals that DenseFusion performs poorly on uncertainty estimation for symmetric objects, for all methods and baselines.  Our analysis revealed that this is due to DenseFusion's lack of robustness to poor segmentation masks.  To demonstrate this,  we evaluated our Learned Comparison method using DenseFusion features but using ground truth masks, instead of estimated masks.  The results, shown in Table~\ref{tbl:masks}, reveal a substantial increase in performance for the log likelihood of symmetric objects, when using ground truth masks instead of estimated masks. This experiment reveals the large contribution of poor segmentation to the overall pose uncertainty in Table~\ref{tbl:loglik}, for DenseFusion on symmetric objects. In contrast, because PoseCNN does not receive as input a segmentation mask, it is more robust to these types of errors.


\begin{table}[h!]
\small
\centering
\begin{tabular}{|c||c|c|c|}
\hline
 & Non-Symmetric & Symmetric & All \\
\hline
Estimated Masks &  1.18 & -5.54 & -0.18 \\
Ground Truth Masks & 1.97 & -0.18 & 1.61 \\
\hline
\end{tabular}
\caption{Mean Log Likelihood of Ground Truth Orientation for Learned Comparison Estimator using DenseFusion Features with estimated and ground truth masks.}
\label{tbl:masks}
\vspace{-2em}
\end{table}







\subsection{Confidence Filtering}
As previously shown~\cite{manhardt2019explaining}, pose uncertainty estimation can be used to robustly filter pose estimates. As we are directly computing the likelihood of an estimate, the output of our algorithm can be used to select which poses to trust and which to reject. Specifically, we use each of our methods to estimate a distribution over orientations. We then compute a pose estimate $\mathbf{\hat{q}}$ from the base pose estimator, and we use our estimated distributions to compute the likelihood at this pose: $p(\mathbf{\hat{q}} \mid \phi(I))$.
For our Learned Comparison method, this requires interpolating the histogram, which we achieve using the interpolation described in Section \ref{sec:multimodal_distribution_regression}.

We test the validity of this process in Table \ref{tbl:error}, which shows the effects of rejecting pose estimates based on likelihood thresholds. In this experiment, we describe these thresholds as multiples of the likelihood of a sample selected at from a uniform distribution, 0.101. As a reminder, this is a probability density, rather than a discrete probability value, and thus ranges from 0 to infinity. For the remaining poses, angular error is calculated with respect to annotated symmetry axes and Average Distance Error (ADD) and Symmetric Average Distance Error (ADD-S) is computed for non-symmetric objects and symmetric objects, respectively. Further details on these evaluation metrics can be found in prior works~\cite{xiang2017posecnn,wang2019densefusion,manhardt2019explaining}.  

Our results can be seen in Table~\ref{tbl:error}, which shows a clear trend of decreasing angular error with an increasing threshold of estimated log likelihood.  This shows that  using a threshold on the estimated log likelihood (using our methods for estimating orientation distributions) is indeed an effective approach for filtering out examples with a large angular error.  Such a threshold can be used to allow a robot to determine when its predictions might be inaccurate.  In such cases, the robot can move its camera to acquire new viewpoints before taking an action, or it can ask a human for help.




\begin{table}[h]
\small
\centering
\begin{tabular}{|c||c|c|c|}
\multicolumn{4}{c}{Learned Comparison (PoseCNN)} \\
\hline
Threshold & Ang Error (deg) & ADD (m)  & Reject (\%) \\
\hline
- & 25.44 & 0.0402 & 0 \\
Uniform & 24.76 & 0.0398 & 3 \\
10x Uniform & 23.69 & 0.0390 & 7 \\
50x Uniform & 17.12 & 0.0374 & 20 \\
100x Uniform & 15.90 & 0.0361 & 34 \\
200x Uniform & 12.72 & 0.0364 & 71 \\
\hline
\multicolumn{4}{l}{} \\
\multicolumn{4}{c}{Bingham Regression (DenseFusion)} \\
\hline
Threshold & Ang Error (deg) & ADD (m)  & Reject (\%) \\
\hline
- & 21.68 & 0.0155 & 0 \\
Uniform & 21.61 & 0.0155 & 0 \\
50x Uniform & 19.08 & 0.0145 & 11 \\
250x Uniform & 16.91 & 0.0135 & 18 \\
1e3x Uniform & 13.74 & 0.0118 & 25 \\
2e3x Uniform & 12.53 & 0.0112 & 30 \\
\hline
\multicolumn{4}{l}{} \\
\multicolumn{4}{c}{(a) Non-Symmetric Objects} \\
\multicolumn{4}{l}{} \\
\multicolumn{4}{c}{Learned Comparison (PoseCNN)} \\
\hline
Threshold & Ang Error (deg) & ADD-S (m)  & Reject (\%) \\
\hline
- & 40.05 & 0.0478 & 0 \\
Uniform & 34.13 & 0.0472 & 13 \\
2x Uniform & 32.60 & 0.0475 & 16 \\
5x Uniform & 29.24 & 0.0468 & 24 \\
15x Uniform & 25.43 & 0.0487 & 40 \\
\hline
\multicolumn{4}{l}{} \\
\multicolumn{4}{c}{(b) Symmetric Objects} \\
\multicolumn{4}{l}{} \\
\end{tabular}
\vspace{-2em}
\caption{Pose error computed on estimates below likelihood thresholds for Non-Symmetric (a) and Symmetric (b) objects. The thresholds are described as multiples of chance, the likelihood of a uniform distribution (0.101).}
\label{tbl:error}
\vspace{-4em}
\end{table}

\section{Conclusion} 
\label{sec:conclusion}

We propose two methods for augmenting existing pose estimation methods with orientation distributions. These methods were compared to a series of uncertainty estimation baselines, evaluated using the log likelihood of the ground-truth orientation. Our findings indicate that, for non-symmetric objects, our learned isotropic Bingham regression gives the best performance. For objects with unknown symmetries, our method for estimating a non-parametric distribution based on a learned feature comparison gives the best performance. We demonstrate that our method can be used to filter out the examples with the worst angular error, for which the robot can choose to capture more information about the environment or request help from a human. Future work will use this uncertainty estimation in the context of tracking or grasping applications; we will also explore how multiple methods for estimating uncertainty can be combined for improved performance.




\bibliographystyle{IEEEtran.bst}
\bibliography{main}

\end{document}


\begin{appendices}

\section{Orientation Gridding}
\label{sec:gridding}
Some of the methods that we describe below represent a discrete multi-modal orientation distribution.  To obtain such a distribution, we need to obtain a uniform sampling of orientations.  Because we are designing an orientation estimation method for use in an object manipulation context, the objects may appear at any arbitrary orientation.  Thus, we want to estimate the object's orientation distribution over the entire SO(3) orientation space. As such, our orientation gridding must cover this entire space.

In \cite{Fox-RSS-19}, a uniform gridding of Euler angles is used, but as shown in \cite{shoemake1992uniform}, this does not produce a uniform sampling of SO(3). Another approach is to use a uniform random sampling of quaternions; however, we desire a deterministic method to ensure uniform coverage of the orientation space and for repeatability.


To generate a uniform grid of orientations over SO(3), we note that an orientation quaternion is a 4-dimensional unit vector, and thus orientations can be represented as points on a unit 3-sphere. By embedding the ``600-cell", a convex regular 4-polytope whose 600 "faces" consist of tetrahedra, into the 3-sphere, we can generate an arbitrarily fine gridding through successive tetrahedral subdivision~\cite{straub2016efficient}. In our experiments, we subdivide each tetrahedra in the 600-cell twice, leading to a set of tetrahedra with 3885 unique vertices. This gridding gives us a maximum distance from any orientation to its closest vertex of about 10 degrees.

\section{Normalizing Histogram Distribution}

Given an unnormalized distribution with a value of $f_i$ at vertex $x_i$ over the half 3-sphere, whose area is $\pi^2$, the normalized distribution is given by $\frac{1}{\eta} f_i$ where 

\begin{align*}
\eta &= \frac{\pi^2}{N} \sum_i^N f_i
\end{align*}

If you look at single nearest neighbor interpolation, this breaks the half sphere into $N$ equi-area regions, each of which has a uniform value of $\frac{1}{\eta} f_i$ and surface area of $\frac{\pi^2}{N}$. The surface integral this interpolation over the half 3-sphere is 

\begin{align*}
\iint_\mathcal{S} p_1(x) d\mathcal{S} &=\sum_i^N \frac{\pi^2}{N} \frac{1}{\eta} f_i \\
&= \frac{1}{\frac{\pi^2}{N} \sum_i^N f_i} \frac{\pi^2}{N} \sum_i^N f_i \\
&= 1
\end{align*}

K nearest neighbor smoothing with inverse distance weighting $d(\mathbf{q}_1, \mathbf{q}_2)$ is then a convex combination of the normalized values, 

\begin{align*}
p_K(\mathbf{q}^*) &= \frac{\sum_k \frac{1}{\eta} f_k d(\mathbf{q}^*, \mathbf{q}_k)}{\sum_k d(\mathbf{q}^*, \mathbf{q}_k)} \\
&= \frac{N}{\pi^2 \sum_i^N f_i} \frac{\sum_k f_k d(\mathbf{q}^*, \mathbf{q}_k)}{\sum_k d(\mathbf{q}^*, \mathbf{q}_k)} \\
\end{align*}

\section{K Nearest Neighbor Loss}
Using the normalization described above, log likelihood of $\mathbf{q}^*$ is 

\begin{align*}
    \mathcal{L}_K (\mathbf{q}^*) &= \log\left(p_K(\mathbf{q}^*)\right) \\
    &= \log\left(\frac{N}{\pi^2 \sum_i^N f_i} \frac{\sum_k f_k d(\mathbf{q}^*, \mathbf{q}_k)}{\sum_k d(\mathbf{q}^*, \mathbf{q}_k)}\right) \\
    &= \log\left(\frac{N}{\pi^2}\right) - \log\left(\sum_i^N f_i\right) + \log\left(\sum_k f_k d(\mathbf{q}^*, \mathbf{q}_k)\right) - \log\left(\sum_k d(\mathbf{q}^*, \mathbf{q}_k)\right) \\
\end{align*}

Removing the terms that are constant with respect to the weights leaves you with

$$
\mathcal{L}_K(\mathbf{q}^*) \propto - \log\left(\sum_i^N f_i\right) + \log\left(\sum_k f_k d(\mathbf{q}^*, \mathbf{q}_k)\right)
$$

\section{Bingham Parameterization}

The Quaternion rotation parameter $\mathbf{M}$ of a Bingham distribution can be decomposed using Cayley’s factorization into $\mathbf{q}_L = \begin{bmatrix} a, b, c, d \end{bmatrix}$ and $\mathbf{q}_R = \begin{bmatrix} p, q, r, s \end{bmatrix}$, such that 

\begin{align*}
    \mathbf{M} &= \begin{bmatrix} a & -b & -c & -d \\ b & a & -d & c \\ c & d & a & -b \\ d & -c & b & a \end{bmatrix} \begin{bmatrix} p & -q & -r & -s \\ q & p & s & -r \\ r & -s & p & q \\ s & r & -q & p \end{bmatrix} \\
\end{align*}

To simplify this representation to distribution centered on $\mathbf{\bar{q}}$, the column associated with the largest value in $\mathbf{Z}$ must be correctly aligned to the column of $\mathbf{M}$ containing the mean quaternion $\mathbf{\bar{q}}$. One possible solution to this is to have $\mathbf{q}_L = \mathbf{\bar{q}}$ and $\mathbf{q}_R = \begin{bmatrix} 1, 0, 0, 0 \end{bmatrix}$. This results in the desired effect, with the second matrix in equation [above] becoming the identity.

To rotated that distribution about the mean quaternion, $\mathbf{\bar{q}}$, you can pose multiply but the three dimensional rotation $\mathbf{R}$ defined by quaternion $\mathbf{q}_{rot} = \begin{bmatrix} x, y, z, w \end{bmatrix}$

\begin{align*}
\mathbf{R} = \begin{bmatrix} 1&               0&               0&               0\\ 
                             0& x^2+y^2-z^2-w^2&   2*(y*z - x*w)&   2*(y*w + x*z) \\
                             0&   2*(y*z + x*w)& x^2-y^2+z^2-w^2&   2*(z*w - x*y) \\
                             0&   2*(y*w - x*z)&   2*(z*w + x*y)& x^2-y^2-z^2+w^2 \end{bmatrix}
\end{align*}

\section{Symmetric Angular Error}

For symmetric objects, standard rotational error is not a meaningful metric. However, given the symmetries of the object, a minimum angular distance between the estimated orientation $\hat{\mathbf{q}}$ and the true orientation $\mathbf{q}$ over all possible symmetric rotations can be calculated.

For the case of discrete rotational symmetries, like in a block, we can compute a symmetry-aware angular distance by iterating over each rotation $\mathbf{p}$ in the set of symmetric rotations $\mathbf{\Phi}$, applying it to the quaternion $\mathbf{q}$ and computing the distance to $\mathbf{\hat{\mathbf{q}}}$, as $\theta = \min_{\mathbf{p} \in \mathbf{\Phi}} \cos^{-1} \left(\| \mathbf{q} \mathbf{p} \cdot \mathbf{\hat{q}}\| \right)$.

For continuous rotational symmetries, as in a bowl, we perform the following computation: given the axis of symmetry $\xi$, we let $\mathbf{x}$ be a vector describing this axis of symmetry $\xi$ rotated by the ground truth quaternion $q$, computed as $\mathbf{x} = \mathbf{q} \xi \mathbf{q}^{-1}$.  Similarly, let $\hat{\mathbf{x}}$ be a vector describing $\xi$ rotated by $\hat{\mathbf{q}}$, computed as $\hat{\mathbf{x}} = \hat{\mathbf{q}} \xi \mathbf{\hat{q}}^{-1}$.  Then we can compute a symmetry-aware angular distance between $\mathbf{q}$ and $\hat{\mathbf{q}}$ as $ \theta = \cos^{-1} \left(\mathbf{x} \cdot \mathbf{\hat{x}}\right)$.

\begin{table*}[h]
\small
\centering
\begin{tabular} {||c||c|c|c|c|c|c|c|c|c|c||}
\hline
  &  & \multicolumn{4}{c|}{Histogram} & \multicolumn{5}{c||}{Bingham} \\
\hline
Objects & Uniform & Conf & Reg & Comp & Cosine & Fixed & Mix & Dropout & Reg Iso & Reg Full \\
\hline
\hline
Non-Symmetric-DF 
 & 23.07 & 85.17 & 73.07 & 72.03 & 65.93 & \textbf{87.94} & \textbf{87.94} & 81.14 & 87.93 & 87.93 \\
Symmetric-DF 
 & 51.62 & \textbf{63.37} & 55.70 & 52.76 & 53.93 & 60.34 & 60.34 & 50.90 & 60.35 & 60.35 \\

\hline
\hline
Non-Symmetric-PC 
 & 23.06 & 68.16 & 79.93 & 80.62 & 84.37 & 85.88 & - & \textbf{86.38} & 85.88 & 85.88 \\
Symmetric-PC 
 & 52.18 & 66.18 & 77.33 & 76.14 & \textbf{80.31} & 77.79 & - & 76.84 & 77.79 & 77.79 \\
\hline
All 
 & 27.73 & 67.84 & 79.51 & 79.90 & 83.71 & 84.57 & - & \textbf{84.84} & 84.57 & 84.57 \\
\hline
\end{tabular}
\caption{AUC of Symmetric Angular Error of Distribution Mode}
\end{table*}

\section{Individual Object Results}
\begin{table*}[h]
\small
\centering
\begin{tabular} {||c||c|c|c|c|c|c|c|c|c|c||}
\hline
  &  & \multicolumn{4}{c|}{Histogram} & \multicolumn{5}{c||}{Bingham} \\
\hline
Objects & Uniform & Conf & Reg & Comp & Cosine & Fixed & Mix & Dropout & Reg Iso & Reg Full \\
\hline
master chef can
 & -2.29 & -0.78 & -2.82 & 1.01 & -1.91 & -0.97 & -0.94 & -1.04 & -1.66 & \textbf{1.21} \\
cracker box
 & -2.29 & \textbf{4.03} & -1.00 & 2.26 & -1.91 & 1.13 & 1.68 & 1.14 & 3.75 & 1.81 \\
sugar box
 & -2.29 & 3.77 & 2.64 & 2.24 & -1.80 & 0.04 & 0.97 & 1.95 & \textbf{5.94} & -0.06 \\
tomato soup can
 & -2.29 & 0.90 & -2.24 & 1.61 & -1.73 & -0.27 & -0.02 & 1.16 & \textbf{2.02} & 1.63 \\
mustard bottle
 & -2.29 & 3.85 & -1.21 & 2.15 & -1.73 & 2.38 & 2.41 & 1.12 & \textbf{4.61} & 1.73 \\
tuna fish can
 & -2.29 & -3.12 & -5.86 & \textbf{0.42} & -2.05 & -1.17 & -2.19 & -0.96 & -0.20 & 0.31 \\
pudding box
 & -2.29 & 2.18 & -0.18 & -1.35 & -1.95 & 1.15 & 1.19 & 0.82 & \textbf{2.64} & 0.47 \\
gelatin box
 & -2.29 & 4.65 & 2.48 & 2.08 & -1.58 & 1.33 & 1.92 & 1.25 & \textbf{6.25} & 2.37 \\
potted meat can
 & -2.29 & 1.18 & -1.36 & 0.61 & -1.91 & -0.01 & 0.03 & 0.95 & \textbf{3.28} & -0.91 \\
banana
 & -2.29 & \textbf{2.58} & -1.12 & 1.03 & -1.90 & -0.65 & -0.45 & 0.11 & 0.58 & -0.86 \\
pitcher base
 & -2.29 & 3.34 & 2.13 & 2.28 & -1.79 & 2.82 & 3.11 & 1.76 & \textbf{4.68} & 1.29 \\
bleach cleanser
 & -2.29 & 3.91 & -4.19 & 1.76 & -1.85 & 1.56 & 2.42 & 1.24 & \textbf{4.70} & 1.77 \\
bowl
 & -2.29 & \textbf{1.37} & -8.88 & -0.46 & -2.13 & -2.45 & -2.33 & -8.99 & -2.77 & -0.28 \\
mug
 & -2.29 & \textbf{3.73} & -6.18 & -1.03 & -2.21 & 0.37 & 0.65 & 1.69 & 2.50 & 1.94 \\
power drill
 & -2.29 & 4.31 & 0.40 & 1.91 & -1.94 & 3.14 & 3.25 & 1.22 & \textbf{5.92} & 1.57 \\
wood block
 & -2.29 & \textbf{2.64} & -6.22 & -2.14 & -2.21 & -2.13 & -2.15 & -13.00 & -2.09 & 0.35 \\
scissors
 & -2.29 & \textbf{3.74} & -13.32 & -0.29 & -2.15 & 0.73 & 0.82 & -2.15 & 0.51 & 1.44 \\
large marker
 & -2.29 & -7.59 & -11.48 & 0.21 & -2.03 & -1.87 & -1.55 & -0.73 & -0.29 & \textbf{0.76} \\
large clamp
 & \textbf{-2.29} & -5.64 & -13.67 & -10.45 & -2.32 & -2.45 & -2.33 & -5.66 & -6.67 & -2.85 \\
extra large clamp
 & -2.29 & -5.20 & -11.94 & -7.49 & -2.19 & -2.24 & -2.27 & -9.28 & -2.92 & \textbf{-2.02} \\
foam brick
 & -2.29 & \textbf{-0.26} & -12.89 & -2.34 & -2.29 & -2.31 & -2.18 & -6.06 & -2.28 & -1.98 \\
\hline
All 
 & -2.29 & 0.86 & -3.90 & 0.09 & -1.95 & -0.01 & 0.18 & -0.74 & \textbf{1.71} & 0.57 \\
\hline
\end{tabular}
\caption{Mean Log Likelihood of Ground Truth Orientation using DenseFusion}
\end{table*}

\begin{table*}[h]
\small
\centering
\begin{tabular} {||c||c|c|c|c|c|c|c|c|c|c||}
\hline
  &  & \multicolumn{4}{c|}{Histogram} & \multicolumn{5}{c||}{Bingham} \\
\hline
Objects & Uniform & Conf & Reg & Comp & Cosine & Fixed & Mix & Dropout & Reg Iso & Reg Full \\
\hline
master chef can
 & -2.29 & -4.99 & 1.40 & \textbf{2.32} & -0.83 & -2.29 & - & 0.09 & -1.57 & -1.70 \\
cracker box
 & -2.29 & -6.06 & 0.92 & -0.09 & -1.06 & -2.29 & - & \textbf{3.71} & 0.90 & 1.81 \\
sugar box
 & -2.29 & 0.77 & 3.19 & 2.62 & -0.82 & -2.28 & - & 4.20 & \textbf{4.63} & 3.17 \\
tomato soup can
 & -2.29 & -0.21 & 2.14 & 2.52 & -0.82 & -2.29 & - & \textbf{3.99} & 1.78 & 3.45 \\
mustard bottle
 & -2.29 & -1.18 & 1.28 & 3.02 & -0.79 & -2.29 & - & \textbf{4.81} & 4.39 & 3.39 \\
tuna fish can
 & -2.29 & -4.24 & 0.91 & \textbf{2.64} & -1.12 & -2.29 & - & 1.23 & -0.12 & 0.74 \\
pudding box
 & -2.29 & -1.62 & 2.97 & 3.13 & -0.95 & -2.29 & - & \textbf{4.54} & 4.12 & 2.65 \\
gelatin box
 & -2.29 & -0.56 & 3.11 & 3.53 & -0.93 & -2.28 & - & 5.73 & \textbf{5.88} & 3.92 \\
potted meat can
 & -2.29 & -2.32 & 1.59 & 1.60 & -0.96 & -2.29 & - & \textbf{3.06} & 0.75 & 2.01 \\
banana
 & -2.29 & -2.01 & 2.52 & 2.27 & -1.12 & -2.28 & - & \textbf{3.70} & 3.06 & 2.93 \\
pitcher base
 & -2.29 & -0.64 & 2.74 & 2.35 & -0.75 & -2.29 & - & \textbf{4.88} & 3.86 & 2.96 \\
bleach cleanser
 & -2.29 & -4.31 & 1.80 & 2.29 & -0.88 & -2.28 & - & \textbf{3.38} & 2.86 & 1.78 \\
bowl
 & -2.29 & -9.73 & -1.95 & \textbf{-1.21} & -1.96 & -2.29 & - & -9.62 & -13.05 & -5.21 \\
mug
 & -2.29 & -2.17 & 2.46 & 2.75 & -0.69 & -2.28 & - & \textbf{4.72} & 2.51 & 2.64 \\
power drill
 & -2.29 & -1.44 & 2.53 & 2.43 & -0.84 & -2.29 & - & 4.17 & \textbf{4.24} & 3.41 \\
wood block
 & -2.29 & -5.01 & 1.09 & -0.51 & -1.44 & -2.28 & - & \textbf{4.49} & 0.93 & 0.03 \\
scissors
 & -2.29 & -4.56 & -1.15 & 0.66 & -1.25 & -2.29 & - & \textbf{1.63} & -1.25 & 0.85 \\
large marker
 & -2.29 & -3.37 & 0.65 & \textbf{1.02} & -1.31 & -2.29 & - & -8.13 & -1.69 & 0.39 \\
large clamp
 & -2.29 & -9.34 & -3.36 & -1.63 & \textbf{-1.59} & -2.29 & - & -3.54 & -7.28 & -5.02 \\
extra large clamp
 & -2.29 & -6.59 & 0.15 & \textbf{0.19} & -1.31 & -2.29 & - & -5.03 & -10.13 & -2.56 \\
foam brick
 & -2.29 & -5.75 & 0.14 & \textbf{1.70} & -1.42 & -2.29 & - & -12.06 & -11.84 & -4.10 \\
\hline
All 
 & -2.29 & -3.31 & 1.37 & \textbf{1.74} & -1.02 & -2.29 & - & 1.43 & 0.20 & 1.11 \\
\hline
\end{tabular}
\caption{Mean Log Likelihood of Ground Truth Orientation using PoseCNN}
\end{table*}

\begin{table*}[h]
\small
\centering
\begin{tabular} {||c||c|c|c|c|c|c|c|c|c|c||}
\hline
  &  & \multicolumn{4}{c|}{Histogram} & \multicolumn{5}{c||}{Bingham} \\
\hline
Objects & Uniform & Conf & Reg & Comp & Cosine & Fixed & Mix & Dropout & Reg Iso & Reg Full \\
\hline
master chef can
 & 0.00 & 65.45 & 56.85 & 56.36 & 49.81 & \textbf{68.89} & \textbf{68.89} & 12.75 & 68.87 & 68.87 \\
cracker box
 & 0.00 & 80.56 & 52.27 & 64.08 & 61.75 & 83.78 & 83.78 & 4.40 & \textbf{83.90} & \textbf{83.90} \\
sugar box
 & 5.85 & 87.75 & 80.27 & 80.42 & 76.32 & 91.01 & 91.01 & 10.24 & \textbf{91.04} & \textbf{91.04} \\
tomato soup can
 & 27.78 & 82.05 & 79.41 & 82.80 & 80.37 & 84.48 & 84.48 & 15.47 & \textbf{84.49} & \textbf{84.49} \\
mustard bottle
 & 0.81 & 88.23 & 80.18 & 80.47 & 79.42 & \textbf{91.92} & \textbf{91.92} & 13.79 & 91.76 & 91.76 \\
tuna fish can
 & 50.86 & 77.78 & 72.51 & 74.24 & 70.42 & \textbf{79.94} & \textbf{79.94} & 19.50 & 79.93 & 79.93 \\
pudding box
 & 18.17 & 81.42 & 65.09 & 58.21 & 71.22 & 85.86 & 85.86 & 23.67 & \textbf{86.12} & \textbf{86.12} \\
gelatin box
 & 66.26 & 94.15 & 89.79 & 88.99 & 90.87 & \textbf{95.96} & \textbf{95.96} & 28.95 & 95.95 & 95.95 \\
potted meat can
 & 33.33 & 81.30 & 73.26 & 72.50 & 70.29 & 82.43 & 82.43 & 12.74 & \textbf{82.48} & \textbf{82.48} \\
banana
 & 35.84 & 72.36 & 33.79 & 50.80 & 33.90 & 74.67 & 74.67 & 6.85 & \textbf{74.69} & \textbf{74.69} \\
pitcher base
 & 0.00 & 84.51 & 72.57 & 71.63 & 46.64 & \textbf{89.80} & \textbf{89.80} & 4.97 & 89.73 & 89.73 \\
bleach cleanser
 & 0.00 & 84.41 & 70.55 & 69.58 & 53.11 & \textbf{87.88} & \textbf{87.88} & 6.24 & 87.74 & 87.74 \\
bowl
 & 31.14 & 26.91 & \textbf{31.98} & 18.22 & 11.45 & 4.65 & 4.65 & 1.80 & 4.65 & 4.65 \\
mug
 & 26.32 & 88.24 & 78.18 & 42.42 & 58.63 & 90.33 & 90.33 & 9.03 & \textbf{90.44} & \textbf{90.44} \\
power drill
 & 10.36 & 87.38 & 79.35 & 69.23 & 55.29 & 90.75 & 90.75 & 6.23 & \textbf{90.76} & \textbf{90.76} \\
wood block
 & 0.00 & \textbf{74.19} & 7.68 & 9.52 & 12.60 & 32.79 & 32.79 & 0.00 & 32.79 & 32.79 \\
scissors
 & 40.05 & 80.58 & 64.39 & 44.00 & 29.82 & 82.03 & 82.03 & 10.05 & \textbf{82.10} & \textbf{82.10} \\
large marker
 & 59.87 & 85.64 & 78.42 & 72.10 & 62.25 & 89.33 & 89.33 & 30.28 & \textbf{89.34} & \textbf{89.34} \\
large clamp
 & 15.79 & \textbf{19.55} & 14.81 & 12.86 & 9.71 & 13.03 & 13.03 & 1.43 & 13.02 & 13.02 \\
extra large clamp
 & 15.24 & 21.67 & 9.64 & 17.18 & 15.21 & 23.50 & 23.50 & 0.19 & \textbf{23.59} & \textbf{23.59} \\
foam brick
 & 48.04 & \textbf{55.88} & 49.30 & 54.08 & 49.02 & 51.68 & 51.68 & 19.75 & 51.73 & 51.73 \\
\hline
All 
 & 20.19 & 73.72 & 62.92 & 61.33 & 55.72 & 74.50 & 74.50 & 10.77 & \textbf{74.51} & \textbf{74.51} \\
\hline
\end{tabular}
\caption{AUC of Average Distance Error of Distribution Mode using DenseFusion}
\end{table*}

\begin{table*}[h]
\small
\centering
\begin{tabular} {||c||c|c|c|c|c|c|c|c|c|c||}
\hline
  &  & \multicolumn{4}{c|}{Histogram} & \multicolumn{5}{c||}{Bingham} \\
\hline
Objects & Uniform & Conf & Reg & Comp & Cosine & Fixed & Mix & Dropout & Reg Iso & Reg Full \\
\hline
master chef can
 & 0.00 & 12.02 & 52.17 & 52.62 & \textbf{60.45} & 50.20 & - & 49.90 & 50.20 & 50.20 \\
cracker box
 & 0.00 & 29.15 & 41.51 & 38.05 & 42.96 & 53.01 & - & \textbf{55.11} & 53.01 & 53.01 \\
sugar box
 & 4.04 & 60.35 & 65.23 & 60.20 & 65.96 & \textbf{68.47} & - & 68.12 & \textbf{68.47} & \textbf{68.47} \\
tomato soup can
 & 22.13 & 67.84 & 66.60 & 66.79 & 68.82 & 68.19 & - & \textbf{69.10} & 68.19 & 68.19 \\
mustard bottle
 & 0.47 & 53.49 & 76.64 & 78.30 & 79.34 & \textbf{81.09} & - & 80.00 & \textbf{81.09} & \textbf{81.09} \\
tuna fish can
 & 44.45 & 68.03 & 66.87 & 70.62 & 70.34 & 70.70 & - & \textbf{71.24} & 70.70 & 70.70 \\
pudding box
 & 10.87 & 38.41 & 60.08 & 62.21 & 60.96 & 62.56 & - & \textbf{62.65} & 62.56 & 62.56 \\
gelatin box
 & 59.30 & 68.56 & 73.61 & 74.21 & 73.61 & \textbf{75.19} & - & 75.18 & \textbf{75.19} & \textbf{75.19} \\
potted meat can
 & 25.91 & 57.76 & 57.24 & \textbf{61.33} & 57.87 & 60.52 & - & 60.66 & 60.52 & 60.52 \\
banana
 & 30.27 & 52.46 & 65.11 & 67.60 & 72.62 & 72.39 & - & \textbf{73.42} & 72.39 & 72.39 \\
pitcher base
 & 0.00 & 23.00 & 50.60 & 46.98 & 52.26 & 53.29 & - & \textbf{54.01} & 53.29 & 53.29 \\
bleach cleanser
 & 0.00 & 37.20 & 44.49 & 45.91 & 48.99 & 50.43 & - & \textbf{50.68} & 50.43 & 50.43 \\
bowl
 & \textbf{16.72} & 10.37 & 5.05 & 4.82 & 7.88 & 3.26 & - & 5.53 & 3.26 & 3.26 \\
mug
 & 16.18 & 48.29 & 52.01 & 57.58 & 57.80 & 58.49 & - & \textbf{59.41} & 58.49 & 58.49 \\
power drill
 & 9.19 & 40.20 & 53.39 & 51.46 & 52.75 & \textbf{55.23} & - & 55.08 & \textbf{55.23} & \textbf{55.23} \\
wood block
 & 0.00 & 12.23 & 15.79 & 4.56 & 8.02 & 27.08 & - & \textbf{29.76} & 27.08 & 27.08 \\
scissors
 & 17.03 & 30.88 & 28.39 & 28.17 & 32.75 & 35.74 & - & \textbf{35.97} & 35.74 & 35.74 \\
large marker
 & 43.03 & 54.66 & 57.15 & 58.25 & \textbf{58.40} & 58.09 & - & 58.38 & 58.09 & 58.09 \\
large clamp
 & 17.03 & 26.86 & 29.12 & \textbf{29.70} & 29.20 & 25.24 & - & 25.86 & 25.24 & 25.24 \\
extra large clamp
 & 11.39 & 12.86 & \textbf{20.92} & 16.95 & 20.48 & 18.52 & - & 18.26 & 18.52 & 18.52 \\
foam brick
 & 42.60 & 50.90 & 40.79 & \textbf{55.96} & 55.55 & 40.25 & - & 40.75 & 40.25 & 40.25 \\
\hline
All 
 & 15.87 & 43.19 & 51.49 & 51.67 & 54.06 & 54.34 & - & \textbf{54.81} & 54.34 & 54.34 \\
\hline
\end{tabular}
\caption{AUC of Average Distance Error of Distribution Mode using PoseCNN}
\end{table*}
\begin{table*}[h]
\small
\centering
\begin{tabular} {||c||c|c|c|c|c|c|c|c|c|c||}
\hline
  &  & \multicolumn{4}{c|}{Histogram} & \multicolumn{5}{c||}{Bingham} \\
\hline
Objects & Uniform & Conf & Reg & Comp & Cosine & Fixed & Mix & Dropout & Reg Iso & Reg Full \\
\hline
master chef can
 & 85.00 & 93.89 & 92.88 & 93.27 & 93.09 & 94.39 & 94.39 & 40.20 & \textbf{94.41} & \textbf{94.41} \\
cracker box
 & 70.90 & 90.66 & 85.45 & 86.78 & 86.97 & 91.73 & 91.73 & 22.71 & \textbf{91.78} & \textbf{91.78} \\
sugar box
 & 77.29 & 94.08 & 91.50 & 91.35 & 89.80 & 95.34 & 95.34 & 31.61 & \textbf{95.39} & \textbf{95.39} \\
tomato soup can
 & 88.35 & 95.15 & 93.92 & 94.95 & 94.77 & \textbf{95.70} & \textbf{95.70} & 33.12 & 95.69 & 95.69 \\
mustard bottle
 & 79.79 & 95.12 & 92.27 & 92.38 & 92.20 & \textbf{96.30} & \textbf{96.30} & 41.88 & 96.26 & 96.26 \\
tuna fish can
 & 90.83 & 95.60 & 94.54 & 95.34 & 93.73 & \textbf{96.00} & \textbf{96.00} & 44.76 & 95.99 & 95.99 \\
pudding box
 & 87.00 & 92.54 & 88.75 & 87.66 & 89.27 & 93.90 & 93.90 & 47.31 & \textbf{93.92} & \textbf{93.92} \\
gelatin box
 & 88.84 & 96.69 & 94.66 & 94.70 & 95.34 & \textbf{97.35} & \textbf{97.35} & 52.04 & 97.35 & 97.35 \\
potted meat can
 & 85.13 & 91.27 & 90.16 & 90.01 & 89.75 & 91.60 & 91.60 & 27.29 & \textbf{91.60} & \textbf{91.60} \\
banana
 & 70.34 & 92.73 & 91.08 & 91.62 & 87.47 & 93.18 & 93.18 & 29.76 & \textbf{93.18} & \textbf{93.18} \\
pitcher base
 & 70.68 & 92.90 & 90.73 & 89.77 & 83.33 & \textbf{94.83} & \textbf{94.83} & 22.23 & 94.82 & 94.82 \\
bleach cleanser
 & 68.16 & 93.04 & 89.90 & 89.35 & 82.71 & \textbf{94.72} & \textbf{94.72} & 25.25 & 94.69 & 94.69 \\
bowl
 & 73.64 & 85.96 & 85.87 & 84.30 & 81.34 & 86.83 & 86.83 & 32.91 & \textbf{86.85} & \textbf{86.85} \\
mug
 & 90.20 & 96.09 & 94.66 & 90.46 & 91.28 & 96.54 & 96.54 & 21.70 & \textbf{96.57} & \textbf{96.57} \\
power drill
 & 68.30 & 93.75 & 90.48 & 85.52 & 81.65 & \textbf{94.94} & \textbf{94.94} & 21.94 & 94.94 & 94.94 \\
wood block
 & 74.66 & 89.96 & 78.24 & 76.93 & 75.96 & \textbf{90.42} & \textbf{90.42} & 16.36 & 90.40 & 90.40 \\
scissors
 & 72.39 & \textbf{92.66} & 89.81 & 83.82 & 82.56 & 92.58 & 92.58 & 38.16 & 92.59 & 92.59 \\
large marker
 & 77.02 & 93.64 & 90.74 & 87.88 & 86.59 & 95.58 & 95.58 & 49.34 & \textbf{95.65} & \textbf{95.65} \\
large clamp
 & 42.42 & 47.48 & 47.41 & 45.93 & 45.21 & 47.72 & 47.72 & 18.24 & \textbf{47.85} & \textbf{47.85} \\
extra large clamp
 & 61.12 & 71.20 & 69.69 & 69.03 & 66.74 & 71.36 & 71.36 & 15.62 & \textbf{71.43} & \textbf{71.43} \\
foam brick
 & 90.62 & 91.62 & 91.76 & 92.00 & 91.35 & 92.79 & 92.79 & 57.61 & \textbf{92.82} & \textbf{92.82} \\
\hline
All 
 & 76.75 & 90.02 & 87.84 & 87.07 & 85.35 & 90.89 & 90.89 & 31.08 & \textbf{90.91} & \textbf{90.91} \\
\hline
\end{tabular}
\caption{AUC of Average Symmetric Distance Error of Distribution Mode using DenseFusion}
\end{table*}

\begin{table*}[h]
\small
\centering
\begin{tabular} {||c||c|c|c|c|c|c|c|c|c|c||}
\hline
  &  & \multicolumn{4}{c|}{Histogram} & \multicolumn{5}{c||}{Bingham} \\
\hline
Objects & Uniform & Conf & Reg & Comp & Cosine & Fixed & Mix & Dropout & Reg Iso & Reg Full \\
\hline
master chef can
 & 76.53 & 76.65 & 84.03 & 83.83 & \textbf{84.12} & 83.80 & - & 84.03 & 83.80 & 83.80 \\
cracker box
 & 63.76 & 68.56 & 72.49 & 72.61 & 73.00 & 76.65 & - & \textbf{77.35} & 76.65 & 76.65 \\
sugar box
 & 71.42 & 82.09 & 82.98 & 82.67 & 83.12 & \textbf{84.03} & - & 83.86 & \textbf{84.03} & \textbf{84.03} \\
tomato soup can
 & 78.92 & 83.31 & 83.26 & 83.25 & 83.42 & 83.43 & - & \textbf{83.58} & 83.43 & 83.43 \\
mustard bottle
 & 78.03 & 87.63 & 89.68 & 89.97 & 90.28 & \textbf{90.72} & - & 90.66 & \textbf{90.72} & \textbf{90.72} \\
tuna fish can
 & 85.99 & 87.86 & \textbf{88.42} & 88.40 & 88.21 & 88.12 & - & 88.36 & 88.12 & 88.12 \\
pudding box
 & 75.24 & 76.58 & 78.26 & 78.67 & 78.35 & \textbf{78.94} & - & 78.88 & \textbf{78.94} & \textbf{78.94} \\
gelatin box
 & 82.21 & 84.64 & 85.35 & 85.63 & 85.41 & \textbf{85.96} & - & 85.91 & \textbf{85.96} & \textbf{85.96} \\
potted meat can
 & 79.02 & 80.82 & 80.62 & 80.95 & 80.66 & 81.04 & - & \textbf{81.06} & 81.04 & 81.04 \\
banana
 & 67.58 & 81.61 & 83.55 & 85.22 & 86.39 & 86.39 & - & \textbf{86.94} & 86.39 & 86.39 \\
pitcher base
 & 64.38 & 71.25 & 77.46 & 77.94 & 77.86 & 78.17 & - & \textbf{78.65} & 78.17 & 78.17 \\
bleach cleanser
 & 56.84 & 69.86 & 72.22 & 71.82 & 72.16 & \textbf{73.05} & - & 73.00 & \textbf{73.05} & \textbf{73.05} \\
bowl
 & 71.50 & 69.61 & 72.47 & \textbf{73.30} & 72.64 & 70.51 & - & 70.19 & 70.51 & 70.51 \\
mug
 & 76.65 & 77.80 & 78.33 & 78.44 & 78.51 & 78.32 & - & \textbf{78.57} & 78.32 & 78.32 \\
power drill
 & 58.22 & 68.59 & 72.47 & 71.15 & 71.78 & \textbf{72.95} & - & 72.78 & \textbf{72.95} & \textbf{72.95} \\
wood block
 & 53.44 & 63.35 & \textbf{65.14} & 63.32 & 64.74 & 62.86 & - & 63.94 & 62.86 & 62.86 \\
scissors
 & 49.20 & 55.95 & 55.61 & 54.70 & 56.99 & \textbf{57.92} & - & 57.48 & \textbf{57.92} & \textbf{57.92} \\
large marker
 & 62.19 & 70.13 & 69.95 & 71.13 & \textbf{71.26} & 70.96 & - & 71.21 & 70.96 & 70.96 \\
large clamp
 & 46.49 & 50.72 & \textbf{52.67} & 52.51 & 51.63 & 52.28 & - & 51.81 & 52.28 & 52.28 \\
extra large clamp
 & 47.93 & 50.27 & 53.69 & 53.62 & 54.40 & 53.52 & - & \textbf{54.58} & 53.52 & 53.52 \\
foam brick
 & 85.72 & 86.00 & 86.88 & 86.88 & \textbf{86.92} & 86.74 & - & 86.77 & 86.74 & 86.74 \\
\hline
All 
 & 68.73 & 74.32 & 76.47 & 76.43 & 76.64 & 77.02 & - & \textbf{77.16} & 77.02 & 77.02 \\
\hline
\end{tabular}
\caption{AUC of Average Symmetric Distance Error of Distribution Mode using PoseCNN}
\end{table*}

\end{appendices}

\clearpage 

\bibliographystyle{IEEEtran.bst}
\bibliography{supplement}